\documentclass[letterpaper, 10 pt, conference]{ieeeconf}  

\IEEEoverridecommandlockouts                              

\usepackage{graphics} 
\usepackage{epsfig} 
\usepackage{mathptmx} 
\usepackage{times} 
\usepackage{amsmath} 
\usepackage{amssymb}  
\usepackage{multirow}
\usepackage{graphicx}
\usepackage{caption}
\usepackage{cite}
\usepackage{hyperref} 
\usepackage{color}

\usepackage{booktabs}
\usepackage[normalem]{ulem}
\useunder{\uline}{\ul}{}

\title{\LARGE \bf
Robot Interaction Behavior Generation based on Social Motion Forecasting for Human-Robot Interaction
}

\author{Esteve Valls Mascaro\footnotemark$^{1}$ and Yashuai Yan\footnotemark$^{1}$ and Dongheui Lee\footnotemark$^{1,2}$
\thanks{$^{1}$Esteve Valls Mascaro and Yashuai Yan and Dongheui Lee are with Autonomous Systems, Technische Universität Wien (TU Wien), Vienna, Austria (e-mail: \texttt{\{esteve.valls.mascaro, yashuai.yan, dongheui.lee\}@tuwien.ac.at}).}%
\thanks{$^{2}$Dongheui Lee is also with the Institute of Robotics and Mechatronics (DLR), German Aerospace Center, Wessling, Germany.}\\%
\href{https://evm7.github.io/ECHO/}{\color{blue} evm7.github.io/ECHO}
}

\begin{document}

\thispagestyle{empty}
\pagestyle{empty}

\twocolumn[{%
\renewcommand\twocolumn[1][]{#1}%
\maketitle
\vspace{-0.5cm}
\begin{center}
    \centering
    \captionsetup{type=figure}
    \includegraphics[width=1\textwidth]{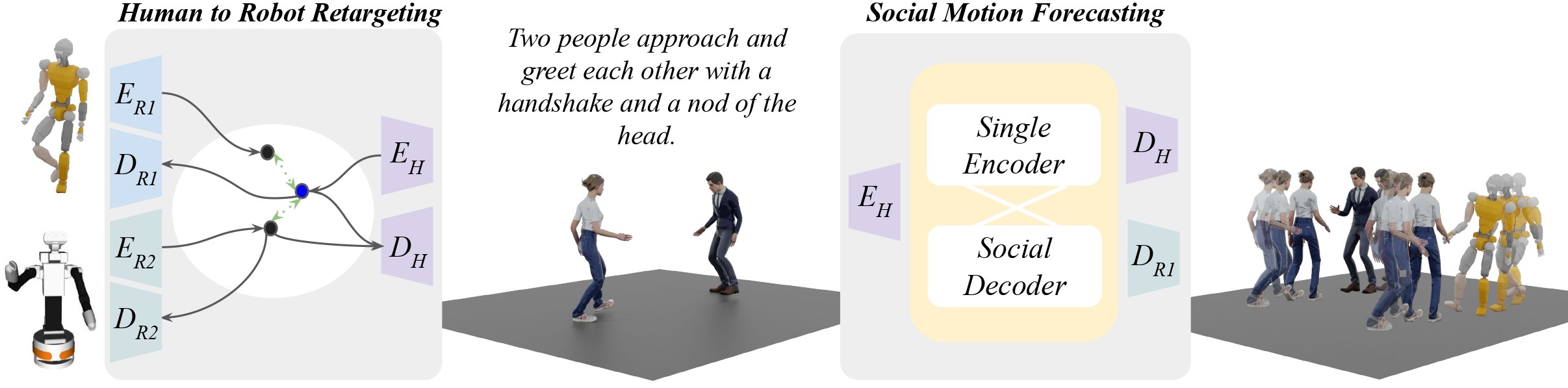}
    \captionof{figure}{\textbf{Overview of our ECHO framework.} First, we learn how to encode ($E$) and decode ($D$) the JVRC-1 robot \cite{jvrc_robot} (in the top left, ${R1}$) and the TIAGo++ robot (in the bottom left, ${R2}$) to a latent representation shared with a human (${H}$)  while preserving its semantics.  Then, we take advantage of this shared space in the social motion forecasting task. Our Single Encoder learns the dynamics of single agents given a textual intention and its past observations. Later, we iteratively refine those motions based on the social context of the surrounding agents using the Social Decoder. Our overall framework can decode the robot's motion in a social environment, closing the gap for natural and accurate Human-Robot Interaction.}
    \label{fig:overview}
\end{center}%
}]

\renewcommand{\thefootnote}{\fnsymbol{footnote}}
\footnotetext{$^{1}$Esteve Valls Mascaro and Yashuai Yan and Dongheui Lee are with Autonomous Systems, Technische Universität Wien (TU Wien), Vienna, Austria (e-mail: \texttt{\{esteve.valls.mascaro, yashuai.yan, dongheui.lee\}@tuwien.ac.at}).}

\footnotetext{$^{2}$Dongheui Lee is also with the Institute of Robotics and Mechatronics (DLR), German Aerospace Center, Wessling, Germany.}
\begin{abstract}

Integrating robots into populated environments is a complex challenge that requires an understanding of human social dynamics. In this work, we propose to model social motion forecasting in a shared human-robot representation space, which facilitates us to synthesize robot motions that interact with humans in social scenarios despite not observing any robot in the motion training.  We develop a transformer-based architecture called ECHO, which operates in the aforementioned shared space to predict the future motions of the agents encountered in social scenarios. Contrary to prior works, we reformulate the social motion problem as the refinement of the predicted individual motions based on the surrounding agents, which facilitates the training while allowing for single-motion forecasting when only one human is in the scene. We evaluate our model in multi-person and human-robot motion forecasting tasks and obtain state-of-the-art performance by a large margin while being efficient and performing in real-time. Additionally, our qualitative results showcase the effectiveness of our approach in generating human-robot interaction behaviors that can be controlled via text commands.

\end{abstract}

\section{INTRODUCTION}

As humans, we commonly find ourselves in social scenarios in which we interact and communicate with each other. Through our experience, we learn how to navigate through those dynamic settings by understanding social norms, individual differences, and the intentions of the surrounding people. While robots have made remarkable strides in various fields, integrating them into social environments still remains a complex challenge. In this work, we propose to tackle this problem by first building a shared representation between humans and robots that we use to learn natural dynamics in social scenarios through motion forecasting. An overview of our framework is depicted in Fig. \ref{fig:overview}.

Human motion forecasting is the task of predicting the human future poses given its past observations. While the community has largely explored individual human motion forecasting \cite{RNN_motion1, sRNN, MSR-GCN, DCT-GCN, mao2020history, 2CH-TR, mascaro2023unified, SIMPLE}, how to effectively model the dependencies in human-human interacting scenarios is still a challenge. When modeling single motions, prior works \cite{RNN_motion1, sRNN, MSR-GCN, DCT-GCN, mao2020history, 2CH-TR, mascaro2023unified, SIMPLE} only consider local skeleton dynamics and do not predict their global trajectory. However, in multi-person forecasting  \cite{TBIFormer, TwoBody, SocialTGCN, d3dpw, mixturedataset,  expi}, there is the need to contextualize the spatial dependencies with the surrounding agents. However, prior works focused on encoding the relationship of multiple humans in scenarios with little or artificially synthesized interactions between the subjects \cite{SocialTGCN, d3dpw, mixturedataset}, or with strong interactions that are not adequate for robots \cite{expi}. On the contrary, we envision scenarios closer to real-world Human-Robot Interactions (HRI) and model highly interactive scenes between humans that are executing a shared action \cite{intergen}, such as handovers, dancing, or greeting.

To effectively model the social dynamics in human-human interactions, we first ground our model to individual motions. We construct a Transformer-based encoder \cite{transformer} that forecasts the next human motion using its own past poses. Previous works  \cite{TBIFormer, TwoBody, expi, SocialTGCN} fuse the spatial relations of multiple humans in the early stage to build a social motion representation. However, we observe that first dealing with individual movements leads to better long-term prediction.  Later on, we refine these generated single motions based on the surrounding agents using a cross-attention decoder \cite{transformer}. By first dealing with individual motions, we ensure that our refinement not only takes into account the current state of the humans in the scene but also the prediction of what other agents and ourselves might do in the future. Our findings observe that understanding other's intentions directly helps to better aggregate all motions into a more natural and human-like social setting. Therefore, we additionally propose to condition the motion synthesis through a text command, that summarizes this overall intention of the social interaction.

Assuming that we are able to model human dynamics in social settings, it is still a challenge to apply this behavior to robots. Previous works \cite{robot_hri, chen2020trust, hri_handover, mascaro2023hoiabot} considered a reactive robot behavior where it first forecasts a human motion and then acts accordingly through a set of learned motions. However, we aim to include the robot behavior in the synthesis of the social dynamics. In our previous work \cite{shared_space_ours} we built a shared latent space between humans and robots that encodes poses with similar semantics. We propose to extend \cite{shared_space_ours} so that it can deal with more complex robot kinematics while having a shared representation for multiple robots and a human decoder. By pre-training our human and robot encoders and decoders, we are able to learn the human dynamics in a human-robot shared space, which facilitates a more efficient and natural human-robot interaction generation.

To sum up, we consider the task of social human motion forecasting but operate in a human-robot shared representation space which allows for the synthesis of accurate and real-time human-robot interactions. The contributions of our paper can be summarized as follows:

\begin{itemize}
    \item A deep-learning architecture that forecasts individual and high-interactive human motions while facilitating their condition on a given interaction.
    \item The encoding of humans and robots in a shared space to synthesize a human-robot interaction in a social context.
    \item An efficient model that achieves state-of-the-art performance in real-time for the social human forecasting task as well as in human-robot collaborative scenarios.
\end{itemize}

\section{RELATED WORK}

\subsection{Human Motion Forecasting}
In the early stages of research, Recurrent Neural Networks (RNNs) \cite{RNN_motion1, RNN_Motion2} were used to understand the time dependencies of human motions and therefore better predict the future human poses. Additionally, \cite{DCT-GCN, MSR-GCN} adopted the Discrete Cosine Transformations (DCT) with Graph Convolutional Networks  (GCN) \cite{gcn} to better model the spatio-temporal relationships. With the success of attention mechanisms, \cite{2CH-TR, mao2020history} used a transformer to model the joint relations in both space and time. Despite significant advances in performance and efficiency, all these models are specialized for motion forecasting tasks within a single human and do not take into consideration the social dynamics essential in interactive scenarios. 

Modeling multi-person interactions has been a long-standing challenge. Early works \cite{Alahi_2014_CVPR, amirian2019social} tackle the global trajectory prediction of humans in a scene. Later, \cite{socialltawarepose} considered the task of multi-person 3D motion forecasting where the context was used to condition the next movements. Recently, \cite{MF_MultiRange} proposed to parallelly leverage individual and multiple human features using transformers to enhance long-term prediction for higher groups of people. On the contrary, \cite{expi} focused on modeling dyadic interactions of humans performing extreme actions via cross-attention. \cite{MF_MultiRange, expi} used a cross-attention mechanism (CA) to enhance one's motion based on others. However, they consider the CA as the synthesis model, while we only leverage CA for motion refinement and address the synthesis with self-attention (SA). To explicitly capture the interactions among joints between the same individual and with others,  \cite{somoformer} operates in each joint with SA, and \cite{TBIFormer} partitions the body into parts and operates in the flattened sequence through SA. While this strategy facilitates better capturing the spatial relationship between joints in each individual, it increases the complexity of the transformer in capturing inter-human dependencies. Finally, \cite{TwoBody} proposes an overall recipe for accurate dyadic motion prediction by reusing DCT and GCN \cite{gcn} in an autoregressive manner. Contrary to our work, all these previous works adopt DCT to encode the temporal dependencies. We show in our experiments that when using DCT the synthesized motions are too smooth and do not capture the nuances between motions. Additionally, we predict the whole future sequence in one step, which avoids the autoregressive approach from \cite{TwoBody, expi} that accumulates error over iterations and collapses in the long term.

\begin{figure*}[]
    \centering
    \includegraphics[width=1\textwidth]{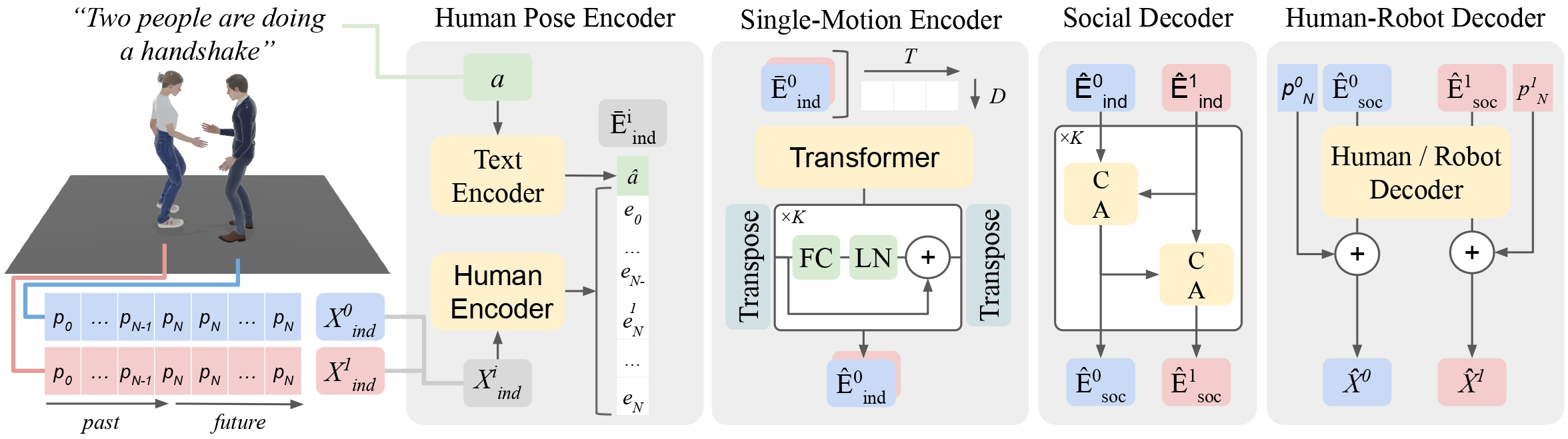}
    \caption{\textbf{Overview of our ECHO architecture}. Our model first focuses on synthesizing individual human motions. First, we pad the observed motion $[\mathbf{p}^i_{t},\cdots,\mathbf{p}^i_{N}]$ for the $i$-th human by repeating the current pose  $\mathbf{p}^i_{N}$ and obtain $\mathbf{X}^i_{ind}$. As our model is conditioned on the social interaction type $a$ and $\mathbf{X}^i_{ind}$, we encode them both and concatenated them to build $\mathbf{\bar{E}}_{ind}^i$. Then, we forecast our individual motions through a self-attention transformer followed by a Temporal MLP with $k$ layers, such that we obtain a single-motion representation  $\mathbf{\hat{E}}_{ind}^i$. As we are considering a social scenario, we iteratively refine the motions per human 0 given the human 1 using cross-attention, and vice versa, obtaining $\mathbf{\hat{E}}_{soc}^0$ and $\mathbf{\hat{E}}_{soc}^1$. This refinement is repeated $K$ times. Finally, we decode each $\mathbf{\hat{E}}_{soc}^i$ and sum the last observed pose $\mathbf{p}^i_{N}$ to make the model invariant to global translations.
    } 
    \label{fig:Architecture}
\end{figure*}

\subsection{Motion retargeting in robotics}
Human motion retargetting has been widely explored in the animation community \cite{rt_animation_contactware, rt_animation_dist_matrix, rt_animation_skeletonaware}. However, in robotics, it is essential to consider not only the naturality of the motions but also the feasibility and adequate control of the robot \cite{MR_Lee, rt_control_1, rt_control_2, rt_deepl1, rt_deepl2, shared_space_ours}. Early works \cite{rt_control_1, rt_control_2} considered optimization-based approaches to transfer a human pose to robots, which required handcrafted features with limited generalizability. To cope with this issue, deep-learning-based methods \cite{rt_deepl1, rt_deepl2, shared_space_ours} learned how to construct a shared latent space to transfer human to robot poses. While \cite{rt_deepl1} and \cite{rt_deepl2} require human annotated or synthetically created human-robot skeleton pairs to learn the retargeting task, \cite{shared_space_ours} proposed to construct a shared representation space that preserves pose semantics without the need of those pairs. In this work, we extend \cite{shared_space_ours} to more complex robots by generating a single latent space from which we can encode or decode different kinematics seemingly. We later use this shared space to predict human-robot interaction motions.

\subsection{Human-Robot Interaction (HRI)}
The rapid growth in robotics is leading to their deployment not only in tightly controlled industrial settings but also in more populated and diverse environments. Therefore, there is a need to develop algorithms that understand the human's intention and assist them in the task at hand \cite{lee_mimetic, lee_imitation, chen2020trust, hri_handover, mascaro2023hoiabot, hri_assistive}.  Early works \cite{nakamura_hri, lee_mimetic, lee_imitation} proposed to recognize this human motion intent through Hidden Markov Models (HMM) and used learned robot primitives for simple HRI. More recently, \cite{motionsyn} focused on the collaborative transportation task through human-robot motion synchronization. However, those works only tackled specific HRI scenarios with close tasks. The success of deep learning opens the doors to properly modeling the dynamic behaviors in HRI \cite{vianello2021human}. For instance, \cite{hri_gestures, Ao2023GestureDiffuCLIP, Zhu_2023_CVPR} proposed to generate natural gestures during the speech, which is essential for embodied conversational robots. While there are clear advances in human motion synthesis conditioned on text \cite{tevet2023human, yuan2023physdiff}, it is still a challenge to generate robot behavior that follows the social dynamics established by humans and generalizes to new scenarios. Our work focuses on that problem and proposes to generate a socially compliant full-body robot behavior while preserving the interaction with other humans and generalizing to a high number of interactive scenarios.

\section{METHODOLOGY}

In this section, we present our ECHO framework for the social motion forecasting task. A visual illustration of our architecture is depicted in Fig. \ref{fig:Architecture}.

\subsection{Problem Formulation}

Let $\mathbf{p}_t^i=[p^i_{t,1},\cdots,p^i_{t,J}]  \in \mathbb{R}^{J\times n}$ be the pose of a $i$-th human at time $t$ composed by $J$ joints and $\mathbf{X^i}=[\mathbf{p}^i_{0}, \cdots, \mathbf{p}^i_{T}]  \in \mathbb{R}^{T \times J \times n}$ a human motion. Each joint $p^i_{t,j}$ is represented as the standard Euclidean $xyz$-position for humans ($n=3$) and joint angle for robot control ($n=1$). Given a social scenario $\mathbf{S}=[\mathbf{X}^0,\cdots,\mathbf{X}^H]$ with $H$ humans at time $N$ where $0\leq N \leq T$, the task of social motion forecasting is defined as the prediction of the subsequent motion $\mathbf{X}_{fut}^i = [\mathbf{p}^i_{N+1}, \cdots, \mathbf{p}^i_{T}]$ per all humans $i$ given their past observations $\mathbf{X}_{past}^i = [\mathbf{p}^i_{0}, \cdots, \mathbf{p}^i_{N}]$. In this work, we only consider dyadic situations ($H=2$). Additionally, we use a global text command $a$ which summarizes the interaction happening in $\mathbf{S}$. 

\subsection{Social Motion Forecasting}
\subsubsection{Motion Forecasting as Refinement}
Similar to prior works \cite{2CH-TR, mascaro2023unified, oreshkin2023motion}, we reformulate the forecasting task by padding the observed motion $\mathbf{X}_{past}^i \in \mathbb{R}^{N \times J \times n}$ to the whole motion dimension $T$ by repeating the last observed pose $\mathbf{X}^i_{ref}=
\mathbf{p}^i_{N}$, obtaining  $\mathbf{X}_{ind}^i \in \mathbb{R}^{T \times J \times n}$. Then, our ECHO network learns a function $f_{\theta}$ to refine the $\mathbf{X}_{ind}^i$ from the reference pose $\mathbf{X}^i_{ref}$, such that $\mathbf{X}_{fut}^i =  f_{\theta}(\mathbf{X}_{ind}^i) + \mathbf{X}^i_{ref}$.

\subsubsection{Pose Encoder}
Given $\mathbf{X}_{ind} \in \mathbb{R}^{ H \times T \times J \times n}$, we initially flatten the joint parameters and encode each body pose $\mathbf{p}_t^i$ in a $D$-representation space $\mathbf{E}=[\mathbf{E}_{ind}^0, \cdots, \mathbf{E}_{ind}^H] \in \mathbb{R}^{ H \times T \times D}$ using a multi-layer perceptron (MLP). In the case of the robot motion generation, we reuse the pre-trained encoder that we will describe in Section \ref{hmr}.

\subsubsection{Single-Motion Encoder}
Our single-motion encoder operates on each individual human independently. We use a transformer model \cite{transformer} to forecast individual human motions conditioned on the observed poses $\mathbf{E}_{ind}^i$ and the overall intention of the interaction $a$. For that, we initially add a sinusoidal positional embedding to $\mathbf{E}_{ind}^i$ to embed the temporal evolution of the motion. Then, we extract the semantic features of the social intention $a$ using \cite{petrovich23tmr}, such that $\hat{a} \in \mathbb{R}^{D}$. We construct the individual motion representation $\mathbf{\bar{E}}_{ind}^i = [\hat{a}, \mathbf{E}_{ind}^i] \in \mathbb{R}^{(T+1) \times D}$ and pass it to a self-attention transformer. Then, inspired by \cite{SIMPLE, mascaro2023unified} in single-motion forecasting, we adopt $k$ temporal MLP layers to iteratively smooth the output of the individual transformer to $\mathbf{\hat{E}}_{ind}^i \in \mathbb{R}^{\times (T+1) \times D}$ by expanding and compressing the time dimensionality of the output.

\subsubsection{Multiple motion forecasting}
Contrary to prior works \cite{MF_MultiRange, expi, somoformer, TBIFormer} that merge multi-person features in the early stage of their architecture, we consider a late-refinement strategy to better preserve the details in a motion. For that, we adopt a series of two cross-attention layers to refine one subject motion based on the others. Our cross-attention mechanism learns how to blend an input \textbf{Q}uery ($\mathbf{Q}$) based on a conditioning \textbf{K}ey ($\mathbf{K}$) and \textbf{V}alue ($\mathbf{V}$). First, we use $\mathbf{\hat{E}}_{ind}^0$ as $\mathbf{Q}$  and $\mathbf{\hat{E}}_{ind}^1$ as $\mathbf{K}$ and $\mathbf{V}$ for the first cross-attention. The goal is that the resulting motion of the subject \textit{0} ($\mathbf{\hat{E}}_{soc}^0$) has been refined to be compliant with subject \textit{1}. This step is now repeated in the inverse order, being $\mathbf{\hat{E}}_{ind}^1$ as $\mathbf{Q}$  and $\mathbf{\hat{E}}_{soc}^0$ as  $\mathbf{K}$ and $\mathbf{V}$. This dual CA is repeated $k$ times to iteratively enhance each motion to be in synchrony with the other subject in the scene.

\subsubsection{Pose Decoder}
Given $\mathbf{\hat{E}}_{soc}^0$ and $\mathbf{\hat{E}}_{soc}^1$, we decode the representation space to the human or robot pose using an MLP layer. In the case of the robot motion generation, we reuse the pre-trained decoder described in Section \ref{hmr}.

\subsection{Losses}
We consider a weighted sum of four different Mean Square Error (MSE) losses to ensure the naturality and dynamism of our generated motion. Here we formulate $\mathbf{X^i}$ and $\mathbf{\hat{X^i}}$ as the predicted and ground-truth motion for the $i$-th human.

\subsubsection{Single ($\mathcal{L}_{ind}$) and Social ($\mathcal{L}_{soc}$) Skeleton Losses} First, we enforce the single-motion encoder to generate plausible motions using $\mathcal{L}_{ind}$. Then, we ensure a proper motion refinement through $\mathcal{L}_{soc}$. $\mathcal{L}_{ind}$ and $\mathcal{L}_{soc}$ are shown in  Eq. \ref{eq_singleloss} and Eq. \ref{eq_pairloss} respectively, where  $D_H$ represents the MLP-based human decoder.

\begin{equation}\label{eq_singleloss}
\small{
{\mathcal{L}_{ind}}(\mathbf{X}^i)=MSE (D_H(\mathbf{\hat{E}}_{ind}^i) - \mathbf{{X}}^i)}
\end{equation}
\begin{equation}\label{eq_pairloss}
\small{
{\mathcal{L}_{soc}}(\mathbf{X}^i)=MSE (D_H(\mathbf{\hat{E}}_{soc}^i) - \mathbf{{X}}^i)}
\end{equation}
\subsubsection{Interaction loss ($\mathcal{L}_{int}$)} To ensure that both agents are in spatial synchrony during the interaction, we force the distance between all joints from agent 0 to 1, referred to as Distance Matrix ($\mathrm{DM}$) to be coherent between the predicted and ground-truth motion. We formulate $\mathcal{L}_{int}$ in Eq. \ref{eq_intloss}.

\begin{equation}\label{eq_intloss}
\small{
{\mathcal{L}_{int}}(\mathbf{X}^0, \mathbf{X}^1)=MSE( \mathrm{DM}(\mathbf{\hat{X}}^0, \mathbf{\hat{X}}^1) - \mathrm{DM}(\mathbf{{X}}^0, \mathbf{{X}}^1))}
\end{equation}

\subsubsection{Bone loss ($\mathcal{L}_{bone}$)} Given the $xyz$-euclidean representations of the body joints, we use $\mathcal{L}_{bone}$ to ensure predicting human poses that are consistent with the bone lengths. We compute the bone lengths from the predictive body joints and ensure their consistency with the real skeleton using MSE.

\subsection{Human-Robot Retargetting}\label{hmr}
The human-to-robot retargeting task aims to find a function $f$ that maps a human pose to a semantically close robot pose ($f: \mathbf{p}_{human} \longmapsto \mathbf{p}_{robot}$). We adopt the strategy from \cite{shared_space_ours} but extend it to construct a meaningful latent space that represents various robots. First, we observe that by considering local joint rotations our model can more effectively capture the nuances in more complex kinematic structures. Second, we learn a unique latent space between various robots and a human, which forces the representation to be more meaningful and close to the semantics of different kinematics. Fig. \ref{fig:overview} presents a simple scheme of our extended proposal for the human to robot retargeting, where all encoders and decoders are MLP layers. We follow similar losses as \cite{shared_space_ours} for the retargeting process, but only consider $\mathcal{L}_{ind}$ and $\mathcal{L}_{soc}$ for the robot agent, as the output from the decoder from \cite{shared_space_ours} is directly joint angles.

\section{EXPERIMENTS}
\begin{table*}[]
\centering
\resizebox{0.97\textwidth}{!}{%
\begin{tabular}{@{}lcccccccccccc@{}}
                      & \multicolumn{4}{c}{\textbf{JPE (mm) $\downarrow$}}                                  & 
                      \multicolumn{4}{c}{\textbf{APJE (mm) $\downarrow$}}                                  &                      
                      \multicolumn{4}{c}{\textbf{FDE (mm)$\downarrow$}}                                  \\ \cmidrule(lr){2-5} \cmidrule(lr){6-9} \cmidrule(lr){10-13}
seconds               & \textbf{0.20}  & \textbf{0.50}  & \textbf{1.00}  & \textbf{1.50}  & \textbf{0.20}  & \textbf{0.50}  & \textbf{1.00}  & \textbf{1.50}  & \textbf{0.20}  & \textbf{0.50}  & \textbf{1.00}  & \textbf{1.50}    \\ \midrule
Zero Velocity & 30.92          & 75.37          & 121.55         & 145.89         & 53.28          & 135.25         & 239.80         & 323.89          & 39.93          & 106.35         & 205.30         & 292.73          \\ \midrule
HisRepIt \cite{mao2020history}           & 39.25          & 68.34          & 100.64         & 120.18         & 56.25          & 106.69         & 179.04         & 234.06          & 38.92          & 78.67          & 147.26         & 202.8          \\ 
SocialTGCN \cite{SocialTGCN}   & 21.56          & 49.13          & 86.65          & 113.65         & 28.77          & 65.60          & 125.48         & 187.88          & 15.59          & 37.38          & 84.05          & 146.24        \\
TBIFormer \cite{TBIFormer}    & 26.55          & 66.27          & 135.47         & 205.67         & {\ul 18.90} &  48.62    &  88.44    & 112.74 & 19.51          & 46.12          & 100.39         & 166.85          \\
ExPI \cite{expi}   & {19.01}    & {56.69} & {127.01} & {203.70} & \textbf{15.30} & \textbf{44.41} & {\ul 85.96} & 113.83    & {13.38} & {37.60} & {93.15} & {166.10} \\
TwoBody \cite{TwoBody}      & \textbf{14.90} & {\ul 38.45}    & {\ul 75.37}    & {\ul 103.43}   & 20.52    & 51.29          & 111.06         & 177.42          & {\ul 11.97}    & {\ul 29.18}    & {\ul 75.34}    & {\ul 140.09}   \\
ECHO (ours)         & {\ul 15.57}    & \textbf{34.37} & \textbf{52.11} & \textbf{70.15} & 20.22         & {\ul 45.01} & \textbf{73.68} & \textbf{110.04}    & \textbf{11.37} & \textbf{25.37} & \textbf{48.85} & \textbf{80.81} \\ \bottomrule
\end{tabular}%
}
\caption{Performance comparison of ECHO model in InterGen dataset \cite{intergen}. All results have been trained specifically for the dataset. A lower score is better. Here. bold indicates the best result and underscores the second-best result.}
\label{tab:socialmotion}
\end{table*}

\subsection{Datasets and Metrics}

\subsubsection{InterGen dataset} InterGen \cite{intergen} is the largest 3D Human motion dataset encompassing $6022$ interactions of two people, accompanied by $16756$ natural language annotations. The dataset contains both daily (e.g. handover, greeting, communications) and more professional (e.g. dancing, boxing) interactions. We adopt the skeleton-based configuration to describe a human with $22$ joints in the $xyz$-euclidean representation. The forecasting task aims to predict the motion of the next $1.5$ sec given an observation of $0.5$ sec.   

\subsubsection{Robot retargetting collection} We randomly sample robot joint angles from the Tiago++ robot and the JVRC-1 \cite{jvrc_robot} humanoid robot to train our human-to-robot imitation network.

\subsubsection{CHICO dataset} CHICO \cite{chicodataset} is the only available 3D motion dataset for Human-Robot Collaboration (HRC). The dataset contains a single operator in a smart factory environment performing seven assembly tasks together with a Kuka LBR robot. The goal is to predict the operator motion in the HRC task. We follow the standard evaluation and predict the next $1000$ ms given $400$ ms in the past.

\subsubsection{Metrics} We use the same evaluation procedure from prior works in multi-person motion forecasting \cite{SocialTGCN, TBIFormer}.  Inspired by the Mean Per Joint Position Error (MPJPE) used in single motion forecasting, we adopt Joint Position Error ({JPE}) to measure millimeter error per global joint position in a given time in the future, and Aligned JPE  ({AJPE}) to only consider the local position error in respect to the root. Equations \ref{eq_jpe} and \ref{eq_ajpe} represent the JPE and AJPE metrics, where $p_{r}$ and $\hat{p}_{r}$ are the estimated and ground-truth root positions of the human body.

\begin{equation}\label{eq_jpe}
\small{
{\rm JPE}(\mathbf{X},\mathbf{\hat{X}})=\frac{1}{H\times J}\sum_{i=1}^{H}\sum_{j=1}^{J}{||X_j^i - \hat{X}_j^i||^2},
}
\end{equation}

\begin{equation}\label{eq_ajpe}
\small{
{\rm AJPE}(\mathbf{X},\mathbf{\hat{X}})={\rm JPE}(\mathbf{X}-p_{r},\mathbf{\hat{X}}-\hat{p}_{r}),
}
\end{equation}

Additionally,  we adopt the Final Displacement Error (FDE) to evaluate the global trajectory of each individual, where $p_{r,t}$ and $\hat{p}_{r,t}$ are the estimated and ground truth root position of the final pose at $t$-th predicted timestamp.
\begin{equation}\label{eq11}
\small{
{\rm FDE}(\mathbf{X},\mathbf{\hat{X}}) = {||p_{r,t} - \hat{p}_{r,t}||^2},
}
\end{equation}

\subsection{Quantitative Evaluation}

\subsubsection{Social Motion Forecasting} 
\begin{figure*}[]
    \centering
    \includegraphics[width=0.95\textwidth]{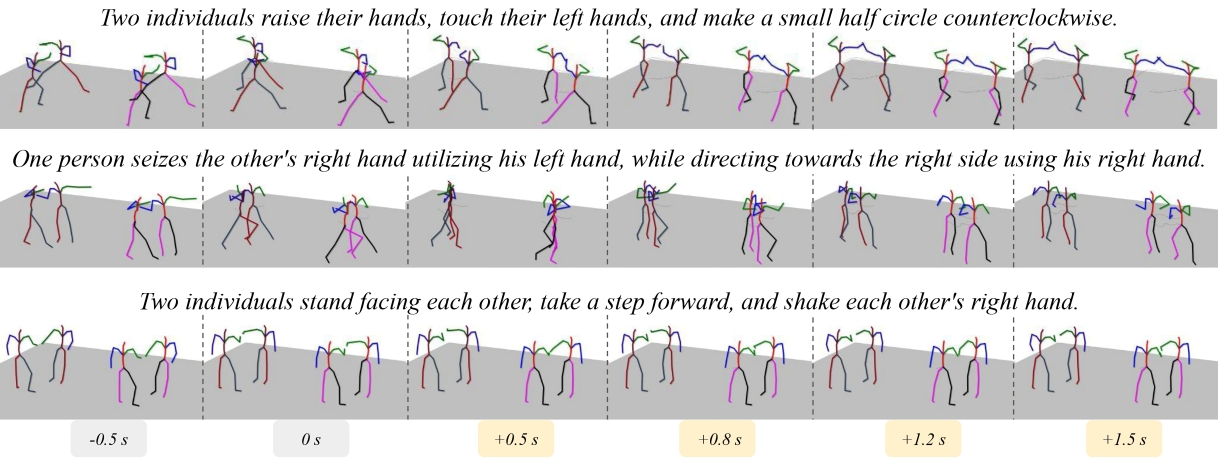}
    \caption{\textbf{Qualitative results for social motion forecasting in the InterGen \cite{intergen} dataset}. Each scenario shows the ground-truth human pair (left) and the predicted (right) per each time horizon.}
    \label{fig:qualitativeresults}
\end{figure*}

Our model is tested in the InterGen dataset for dyadic motion forecasting in social settings and outperforms on average all state-of-the-art models in multi-person motion forecasting, as shown in Table \ref{tab:socialmotion}. All models reported in the paper have been trained for the InterGen dataset with the same training configuration for a fair comparison. We consider \textit{Zero Velocity} as the repetition of the last pose observed, which acts as the simplest baseline for our evaluation. Additionally, \cite{mao2020history} is only focused on single human motion forecasting, which shows the benefit of considering the scene context to refine a given individual motion. On the contrary, \cite{SocialTGCN ,TBIFormer, TwoBody, expi} are multi-person motion forecasting models. Table \ref{tab:socialmotion} demonstrates that the use of auto-regressive approaches \cite{TwoBody, expi} facilitates a better prediction in the short-term, but fails to capture the long-term dependencies of a model. Moreover, as our model predicts the whole motion in one shot, the inference is much faster, which is crucial for real-world HRI. Our model outperforms the other baselines in all metrics on average. We showcase the performance of our ECHO model in Fig. \ref{fig:qualitativeresults} for the social motion forecasting with human decoding. All models were trained in a single GPU for 150 epochs using an exponential decay scheduler and AdamW as an optimizer, with 5 epochs warmup.

\subsubsection{Motion Forecasting in Human-Robot Collaboration} 
Additionally, we train and evaluate our model in the CHICO dataset \cite{chicodataset} for the single motion forecasting conditioned by the robot motion and the observed human motion. Due to the different number of joints between the human operator and the robot, we use different MLPs to encode each agent. Similar to the original work \cite{chicodataset}, we only consider the MPJPE metrics for the short-term (400ms) and long-term (1000ms) horizons. Our results, reported in Table \ref{tab:chicodataset}, show that our ECHO model outperforms previous baselines, mostly in long-term forecasting.

\begin{table}[]
\centering
\caption{Quantitative evaluation of the short (400ms) and long-term (1000ms) motion forecasting in the CHICO dataset \cite{chicodataset} reported in MPJPE. Here, bold indicates the best result and underscores the second-best result.}
\resizebox{0.27\textwidth}{!}{%
\begin{tabular}{lrr}
milliseconds (ms) & \textbf{400}    & \textbf{1000}    \\ \midrule
Zero Velocity     & 162.0    & 282.0    \\ \midrule
HisRepIt \cite{mao2020history}      & 54.6   & 91.6   \\
MRS-GCN \cite{MSR-GCN}           & 54.1   & 90.7   \\
STS-GCN \cite{stsgcn}            & 53.0     & 87.4   \\
SeS-GCN \cite{chicodataset}          & {\ul 48.8}   & {\ul 85.3}   \\
ECHO (ours)              & \textbf{47.1} & \textbf{80.5} \\ \bottomrule
\end{tabular}%
}
\label{tab:chicodataset}
\end{table}

\subsubsection{Human to Robot Motion} 
\begin{figure}[]
    \centering
    \includegraphics[width=0.46\textwidth]{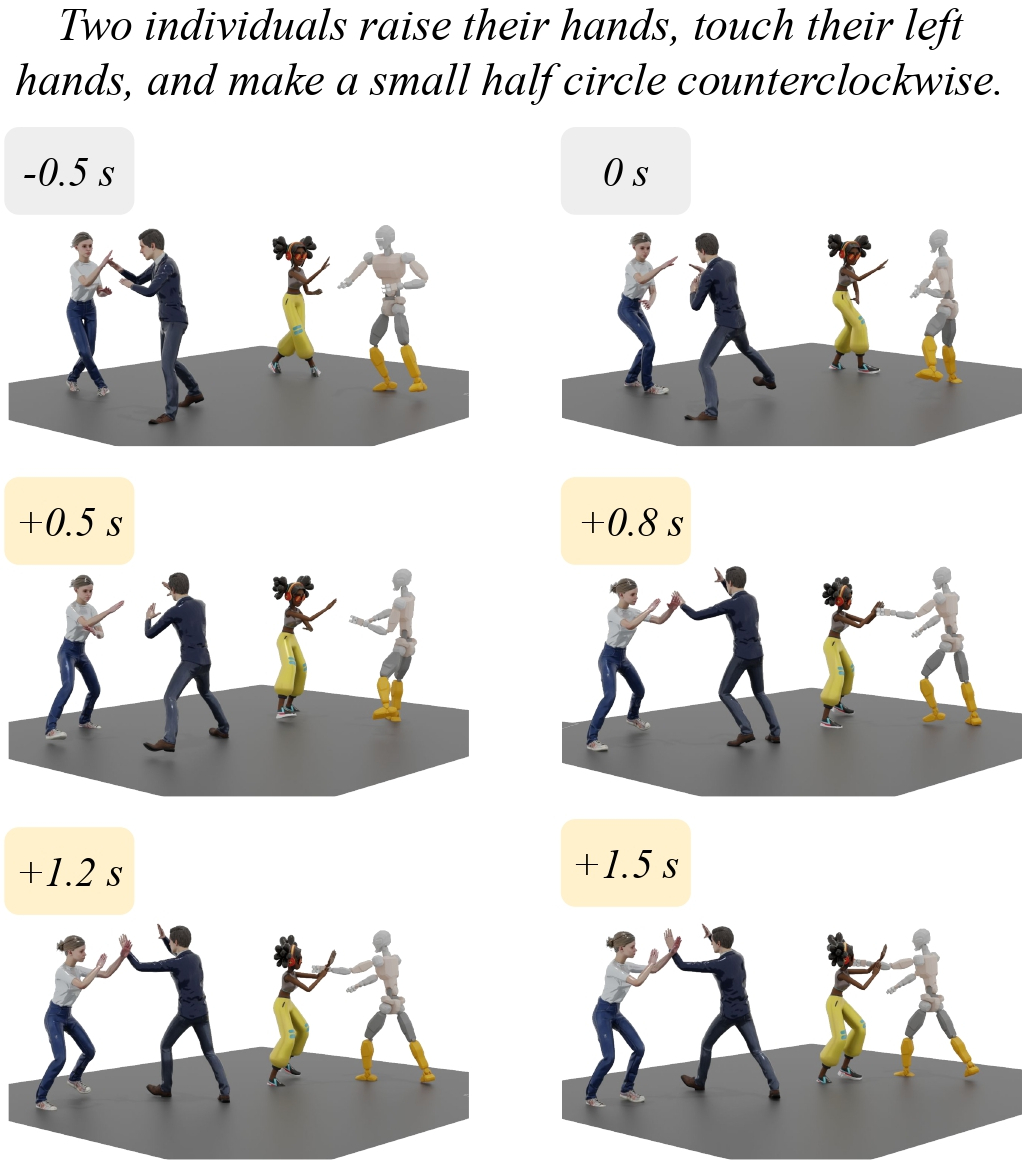}
    \caption{\textbf{Social motion forecasting for Human-Robot Interaction}. Human-Human pair represents the ground truth, while the human-robot pair represents the forecasted human-robot interaction.}
    \label{fig:human2robot}
\end{figure}

We train the human-to-robot retargeting for only the TIAGo++ robot as \cite{shared_space_ours} and also use our new approach for various robots. Our results have lower reconstruction error in the joint angle ($0.005$ versus $0.009$) while additionally decoding into the JVRC-1  robot. Additionally, we showcase in Figure \ref{fig:human2robot} the qualitative evaluation to demonstrate the effectiveness of our overall ECHO framework in decoding HRI.

\subsection{Ablation Study}

\begin{table*}[]
\centering
\resizebox{0.97\textwidth}{!}{%
\begin{tabular}{@{}lcccccccccccc@{}}
                      & \multicolumn{4}{c}{\textbf{JPE (mm) $\downarrow$}}                                  & 
                      \multicolumn{4}{c}{\textbf{APJE (mm) $\downarrow$}}                                  &                      
                      \multicolumn{4}{c}{\textbf{FDE (mm) $\downarrow$}}                                  \\ \cmidrule(lr){2-5} \cmidrule(lr){6-9} \cmidrule(lr){10-13}
seconds               & \textbf{0.20}  & \textbf{0.50}  & \textbf{1.00}  & \textbf{1.50}  & \textbf{0.20}  & \textbf{0.50}  & \textbf{1.00}  & \textbf{1.50}  & \textbf{0.20}  & \textbf{0.50}  & \textbf{1.00}  & \textbf{1.50}    \\ \midrule

w/ DCT & 17.88                           & 36.52                              & 54.01                & 72.25                        & 23.40                              & 47.98                & 75.78                & 111.63               & 13.22                              & 27.05                      & 48.80                      & 81.10                     \\
w/o TempMLP &17.77 & 36.59 &  53.95 & 72.09 &22.63 & 47.44 &75.26 &111.07 & 12.17 & 26.09 & {\ul 47.94} & {\ul 80.24} \\
w/o Text &19.64 & 37.39 & 55.13 & 73.33 & 24.53 & 48.86 & 76.56 & 112.43 & 12.93 &26.78  & 48.56 & 80.83 \\
w/o Baseline & 18.41                           & 39.55                              & 61.35                      & 80.69                              & 23.96                              & 51.91                & 85.79                & 126.21               & 13.60                               & 29.12                      & 55.13                      & 93.09                       \\ \midrule

w/o  $\mathcal{L}_{ind}$ &  16.99                           & 36.97                              & 54.10                      & 71.19                     & 21.57                        & 47.95                & 75.52          & \textbf{109.33}      & 11.95                        & 26.72                      & 48.42                & \textbf{79.09}             \\  \midrule

w/o Iterative Refinement & \textbf{15.46}       & {\ul 34.47}          & {\ul 52.71}          & 70.68                & \textbf{20.17}       & {\ul 45.25}          & {\ul 74.56}          & 111,19               & {\ul 11.39}          & {\ul 25.50}          & 48.29                & 81.74                \\ \midrule

ECHO (ours)         & {\ul 15.57}          & \textbf{34.37}       & \textbf{52.11}       & \textbf{70.15}       & {\ul 20.22}          & \textbf{45.01}       & \textbf{73.68}       & {\ul 110.04}         & \textbf{11.37}       & \textbf{25.37}       & \textbf{47.85}       & 80.81     \\ \bottomrule
\end{tabular}%
}
\caption{Ablation study of our ECHO model for the social motion forecasting task in the InterGen dataset \cite{intergen}.}
\label{tab:abl_study}
\end{table*}

This section provides a systematic assessment of the approaches proposed.  The results of the ablation study are presented in Table \ref{tab:abl_study}.

\subsubsection{Discrete Cosine Transformation (DCT)} Contrary to prior models, we decided not to adopt the DCT to avoid over-smooth motion generations. We observed that DCT is beneficial for models when they are not trained in large-scale datasets, as it facilitates generalization but fails to capture the nuances of different motions.

\subsubsection{Text conditioning} While our works benefit from text to condition the motion forecasting to a known social goal, we also asses the performance without this feature for a fair comparison with the baseline models. We observe that using text improves the results, mostly in short-term horizons, but do not suppose a large improvement. Our ECHO model still outperforms by large margins previous baselines despite not using text as a condition.
\subsubsection{Temporal Multi-Layer Perceptron (TempMLP)} The use of TempMLP has also been adopted by prior works in the motion forecasting field, such as \cite{SIMPLE, mascaro2023unified, TBIFormer}. We observe that TempMLP improves the flow in the short-term horizons as it smoothes out the transition from observed to predicted poses. However, it causes the model to use fixed-length inputs, which tends to reduce the long-term overall performance.

\subsubsection{Baseline approach} Thanks to our refinement strategy, our ECHO model only needs to learn the variation of the future poses with respect to the last pose observed. Our results reinforce the benefit of this strategy.

\subsubsection{Individual loss ($\mathcal{L}_{ind}$)} The use of $\mathcal{L}_{ind}$ benefit the model in the short-term, as each human focus more on its own motion. However, it slightly reduces the importance of social motion, which is key for better long-term performance.

\subsubsection{Iterative Refinement}. We assess the iterative refinement of the individual motions by only considering one dual CA ($k=1$). For a fair evaluation, we extend the number of layers of each CA to $k$, so the model has the same number of parameters. We observe our iterative refinement improves the forecasting in the long term.

\section{CONCLUSIONS}
In this paper, we propose a two-step framework that first learns how humans behave in social scenarios to generate a natural Human-Robot Interaction (HRI). First, we build a unique representation space shared between humans and various robot skeletons that preserves the semantics of the poses while facilitating its pose retargeting. Then, we develop a single-to-social transformer architecture, called ECHO, that learns how humans behave in social scenarios through the motion forecasting task. ECHO understands the current scene and synthesizes a natural and meaningful robot motion to interact with a human. Our results support the approaches taken for our framework, which outperforms the state-of-the-art by large margins in the largest dyadic human motion dataset available, as well as in the field of motion forecasting for Human-Robot Collaborative tasks. In conclusion, our approach can decode a compliant robot motion in a social environment, leading to a more natural and accurate Human-Robot Interaction.

\section*{ACKNOWLEDGMENT}
This work is funded by Marie Sklodowska-Curie Action Horizon 2020 (Grant agreement No. 955778) for project 'Personalized Robotics as Service Oriented Applications' (PERSEO).

\bibliographystyle{ieeetr}
\bibliography{relatedworks}

\begin{thebibliography}{10}

\bibitem{jvrc_robot}
M.~Okugawa, K.~Oogane, M.~Shimizu, Y.~Ohtsubo, T.~Kimura, T.~Takahashi, and
  S.~Tadokoro, ``Proposal of inspection and rescue tasks for tunnel disasters
  — task development of japan virtual robotics challenge,'' in {\em 2015 IEEE
  International Symposium on Safety, Security, and Rescue Robotics (SSRR)},
  pp.~1--2, 2015.

\bibitem{RNN_motion1}
K.~Fragkiadaki, S.~Levine, P.~Felsen, and J.~Malik, ``Recurrent network models
  for human dynamics,'' in {\em IEEE international conference on computer
  vision}, pp.~4346--4354, 2015.

\bibitem{sRNN}
A.~Jain, A.~R. Zamir, S.~Savarese, and A.~Saxena, ``Structural-rnn: Deep
  learning on spatio-temporal graphs,'' in {\em Conference on Computer Vision
  and Pattern Recognition (CVPR)}, pp.~5308--5317, 2016.

\bibitem{MSR-GCN}
L.~Dang, Y.~Nie, C.~Long, Q.~Zhang, and G.~Li, ``Msr-gcn: Multi-scale residual
  graph convolution networks for human motion prediction,'' in {\em IEEE/CVF
  International Conference on Computer Vision (ICCV)}, pp.~11467--11476,
  October 2021.

\bibitem{DCT-GCN}
W.~Mao, M.~Liu, M.~Salzmann, and H.~Li, ``Learning trajectory dependencies for
  human motion prediction,'' in {\em International Conference on Computer
  Vision (ICCV)}, pp.~9489--9497, 2019.

\bibitem{mao2020history}
W.~Mao, M.~Liu, and M.~Salzmann, ``History repeats itself: Human motion
  prediction via motion attention,'' in {\em European Conference on Computer
  Vision (ECCV)}, pp.~474--489, 2020.

\bibitem{2CH-TR}
E.~Valls~Mascaro, S.~Ma, H.~Ahn, and D.~Lee, ``Robust human motion forcasting
  using transformer-based model,'' in {\em 2022 IEEE/RSJ International
  Conference on Intelligent Robots and Systems (IROS)}, pp.~10674--10680, 2022.

\bibitem{mascaro2023unified}
E.~V. Mascaro, H.~Ahn, and D.~Lee, ``A unified masked autoencoder with
  patchified skeletons for motion synthesis,'' {\em arXiv preprint
  arXiv:2308.07301}, 2023.

\bibitem{SIMPLE}
W.~Guo, Y.~Du, X.~Shen, V.~Lepetit, X.~Alameda-Pineda, and F.~Moreno-Noguer,
  ``Back to mlp: A simple baseline for human motion prediction,'' in {\em
  IEEE/CVF Winter Conference on Applications of Computer Vision (WACV)},
  pp.~4809--4819, January 2023.

\bibitem{TBIFormer}
X.~Peng, S.~Mao, and Z.~Wu, ``Trajectory-aware body interaction transformer for
  multi-person pose forecasting,'' in {\em IEEE/CVF Conference on Computer
  Vision and Pattern Recognition (CVPR)}, pp.~17121--17130, June 2023.

\bibitem{TwoBody}
M.~R.~U. Rahman, L.~Scofano, E.~De~Matteis, A.~Flaborea, A.~Sampieri, and
  F.~Galasso, ``Best practices for 2-body pose forecasting,'' in {\em IEEE/CVF
  Conference on Computer Vision and Pattern Recognition}, 2023.

\bibitem{SocialTGCN}
X.~Peng, X.~Zhou, Y.~Luo, H.~Wen, and Z.~Wu, ``The mi-motion dataset and
  benchmark for 3d multi-person motion prediction,'' 2023.
\newblock arXiv preprint arXiv:2306.13566, 2023.

\bibitem{d3dpw}
T.~von Marcard, R.~Henschel, M.~Black, B.~Rosenhahn, and G.~Pons-Moll,
  ``Recovering accurate 3d human pose in the wild using imus and a moving
  camera,'' in {\em European Conference on Computer Vision (ECCV)}, sep 2018.

\bibitem{mixturedataset}
J.~Wang, H.~Xu, M.~Narasimhan, and X.~Wang, ``Multi-person 3d motion prediction
  with multi-range transformers,'' {\em Advances in Neural Information
  Processing Systems}, vol.~34, 2021.

\bibitem{expi}
W.~Guo, X.~Bie, X.~Alameda-Pineda, and F.~Moreno–Noguer, ``Multi-person
  extreme motion prediction,'' in {\em 2022 IEEE/CVF Conference on Computer
  Vision and Pattern Recognition (CVPR)}, pp.~13043--13054, 2022.

\bibitem{intergen}
H.~Liang, W.~Zhang, W.~Li, J.~Yu, and L.~Xu, ``Intergen: Diffusion-based
  multi-human motion generation under complex interactions,'' {\em arXiv
  preprint arXiv:2304.05684}, 2023.

\bibitem{transformer}
A.~Vaswani, N.~Shazeer, N.~Parmar, J.~Uszkoreit, L.~Jones, A.~N. Gomez,
  {\L}.~Kaiser, and I.~Polosukhin, ``Attention is all you need,'' {\em Advances
  in neural information processing systems (NeurIPS)}, vol.~30, 2017.

\bibitem{robot_hri}
T.~Kopp, M.~Baumgartner, and S.~Kinkel, ``Success factors for introducing
  industrial human-robot interaction in practice: an empirically driven
  framework,'' {\em The International Journal of Advanced Manufacturing
  Technology}, vol.~112, 01 2021.

\bibitem{chen2020trust}
M.~Chen, S.~Nikolaidis, H.~Soh, D.~Hsu, and S.~Srinivasa, ``Trust-aware
  decision making for human-robot collaboration: Model learning and planning,''
  {\em ACM Transactions on Human-Robot Interaction (THRI)}, vol.~9, no.~2,
  pp.~1--23, 2020.

\bibitem{hri_handover}
J.~Zhang, H.~Liu, Q.~Chang, L.~Wang, and R.~X. Gao, ``Recurrent neural network
  for motion trajectory prediction in human-robot collaborative assembly,''
  {\em CIRP Annals}, vol.~69, no.~1, pp.~9--12, 2020.

\bibitem{mascaro2023hoiabot}
E.~V. Mascaro, D.~Sliwowski, and D.~Lee, ``{HOI}4{ABOT}: Human-object
  interaction anticipation for human intention reading assistive ro{BOT}s,'' in
  {\em 7th Annual Conference on Robot Learning}, 2023.

\bibitem{shared_space_ours}
Y.~Yan, E.~V. Mascaro, and D.~Lee, ``Unsupervised human-to-robot motion
  retargeting via expressive latent space,'' 2023.
\newblock arXiv preprint arXiv:2309.05310, 2023.

\bibitem{RNN_Motion2}
J.~Martinez, M.~J. Black, and J.~Romero, ``On human motion prediction using
  recurrent neural networks,'' in {\em Conference on Computer Vision and
  Pattern Recognition (CVPR)}, pp.~2891--2900, 2017.

\bibitem{gcn}
T.~N. Kipf and M.~Welling, ``Semi-supervised classification with graph
  convolutional networks,'' {\em arXiv preprint arXiv:1609.02907}, 2016.

\bibitem{Alahi_2014_CVPR}
A.~Alahi, V.~Ramanathan, and L.~Fei-Fei, ``Socially-aware large-scale crowd
  forecasting,'' in {\em IEEE Conference on Computer Vision and Pattern
  Recognition (CVPR)}, June 2014.

\bibitem{amirian2019social}
J.~Amirian, J.-B. Hayet, and J.~Pettr{\'e}, ``Social ways: Learning multi-modal
  distributions of pedestrian trajectories with gans,'' in {\em IEEE/CVF
  Conference on Computer Vision and Pattern Recognition Workshops}, 2019.

\bibitem{socialltawarepose}
V.~Adeli, E.~Adeli, I.~Reid, J.~C. Niebles, and H.~Rezatofighi, ``Socially and
  contextually aware human motion and pose forecasting,'' {\em IEEE Robotics
  and Automation Letters}, vol.~5, no.~4, pp.~6033--6040, 2020.

\bibitem{MF_MultiRange}
J.~Wang, H.~Xu, M.~Narasimhan, and X.~Wang, ``Multi-person 3d motion prediction
  with multi-range transformers,'' in {\em Advances in Neural Information
  Processing Systems} (M.~Ranzato, A.~Beygelzimer, Y.~Dauphin, P.~Liang, and
  J.~W. Vaughan, eds.), vol.~34, Curran Associates, Inc., 2021.

\bibitem{somoformer}
E.~Vendrow, S.~Kumar, E.~Adeli, and H.~Rezatofighi, ``Somoformer: Multi-person
  pose forecasting with transformers,'' 2022.
\newblock arXiv preprint arXiv:2208.14023, 2022.

\bibitem{rt_animation_contactware}
R.~Villegas, D.~Ceylan, A.~Hertzmann, J.~Yang, and J.~Saito, ``Contact-aware
  retargeting of skinned motion,'' 2021.
\newblock arXiv preprint arXiv:2109.07431, 2021.

\bibitem{rt_animation_dist_matrix}
J.~Zhang, J.~Weng, D.~Kang, F.~Zhao, S.~Huang, X.~Zhe, L.~Bao, Y.~Shan,
  J.~Wang, and Z.~Tu, ``Skinned motion retargeting with residual perception of
  motion semantics \& geometry,'' in {\em IEEE/CVF Conference on Computer
  Vision and Pattern Recognition}, 2023.

\bibitem{rt_animation_skeletonaware}
K.~Aberman, P.~Li, D.~Lischinski, O.~Sorkine-Hornung, D.~Cohen-Or, and B.~Chen,
  ``Skeleton-aware networks for deep motion retargeting,'' {\em ACM
  Transactions on Graphics}, vol.~39, no.~4, 2020.

\bibitem{MR_Lee}
C.~Ott, D.~Lee, and Y.~Nakamura, ``Motion capture based human motion
  recognition and imitation by direct marker control,'' in {\em Humanoids 2008
  - 8th IEEE-RAS International Conference on Humanoid Robots}, 2008.

\bibitem{rt_control_1}
W.~Gomes, V.~Radhakrishnan, L.~Penco, V.~Modugno, J.-B. Mouret, and S.~Ivaldi,
  ``Humanoid whole-body movement optimization from retargeted human motions,''
  in {\em 2019 IEEE-RAS 19th International Conference on Humanoid Robots
  (Humanoids)}, 2019.

\bibitem{rt_control_2}
S.~Choi and J.~Kim, ``Towards a natural motion generator: a pipeline to control
  a humanoid based on motion data,'' in {\em 2019 IEEE/RSJ International
  Conference on Intelligent Robots and Systems (IROS)}, 2019.

\bibitem{rt_deepl1}
S.~Choi, M.~Pan, and J.~Kim, ``Nonparametric motion retargeting for humanoid
  robots on shared latent space,'' 07 2020.

\bibitem{rt_deepl2}
S.~Choi, M.~J. Song, H.~Ahn, and J.~Kim, ``Self-supervised motion retargeting
  with safety guarantee,'' in {\em 2021 IEEE International Conference on
  Robotics and Automation (ICRA)}, pp.~8097--8103, IEEE, 2021.

\bibitem{lee_mimetic}
D.~Lee, C.~Ott, and Y.~Nakamura, ``Mimetic communication model with compliant
  physical contact in human—humanoid interaction,'' {\em The International
  Journal of Robotics Research}, vol.~29, no.~13, pp.~1684--1704, 2010.

\bibitem{lee_imitation}
J.~Medina~Hernández, M.~Lawitzky, A.~Mörtl, D.~Lee, and S.~Hirche, ``An
  experience-driven robotic assistant acquiring human knowledge to improve
  haptic cooperation,'' pp.~2416--2422, 09 2011.

\bibitem{hri_assistive}
A.~Jevtić, A.~Flores~Valle, G.~Alenyà, G.~Chance, P.~Caleb-Solly,
  S.~Dogramadzi, and C.~Torras, ``Personalized robot assistant for support in
  dressing,'' {\em IEEE Transactions on Cognitive and Developmental Systems},
  vol.~11, no.~3, pp.~363--374, 2019.

\bibitem{nakamura_hri}
Y.~Nakamura, W.~Takano, and K.~Yamane, ``Mimetic communication theory for
  humanoid robots interacting with humans,'' in {\em Robotics Research}
  (S.~Thrun, R.~Brooks, and H.~Durrant-Whyte, eds.), (Berlin, Heidelberg),
  pp.~128--139, Springer Berlin Heidelberg, 2007.

\bibitem{motionsyn}
L.~Yang, Y.~Li, and D.~Huang, ``Motion synchronization in human-robot
  co-transport without force sensing,'' in {\em 2018 37th Chinese Control
  Conference (CCC)}, pp.~5369--5374, 2018.

\bibitem{vianello2021human}
L.~Vianello, L.~Penco, W.~Gomes, Y.~You, S.~M. Anzalone, P.~Maurice, V.~Thomas,
  and S.~Ivaldi, ``Human-humanoid interaction and cooperation: a review,'' {\em
  Current Robotics Reports}, vol.~2, no.~4, pp.~441--454, 2021.

\bibitem{hri_gestures}
P.~J. Yazdian, M.~Chen, and A.~Lim, ``Gesture2vec: Clustering gestures using
  representation learning methods for co-speech gesture generation,'' in {\em
  2022 IEEE/RSJ International Conference on Intelligent Robots and Systems
  (IROS)}, pp.~3100--3107, 2022.

\bibitem{Ao2023GestureDiffuCLIP}
T.~Ao, Z.~Zhang, and L.~Liu, ``Gesturediffuclip: Gesture diffusion model with
  clip latents,'' {\em ACM Trans. Graph.}

\bibitem{Zhu_2023_CVPR}
L.~Zhu, X.~Liu, X.~Liu, R.~Qian, Z.~Liu, and L.~Yu, ``Taming diffusion models
  for audio-driven co-speech gesture generation,'' in {\em IEEE/CVF Conference
  on Computer Vision and Pattern Recognition (CVPR)}, pp.~10544--10553, June
  2023.

\bibitem{tevet2023human}
G.~Tevet, S.~Raab, B.~Gordon, Y.~Shafir, D.~Cohen-or, and A.~H. Bermano,
  ``Human motion diffusion model,'' in {\em The Eleventh International
  Conference on Learning Representations}, 2023.

\bibitem{yuan2023physdiff}
Y.~Yuan, J.~Song, U.~Iqbal, A.~Vahdat, and J.~Kautz, ``Physdiff: Physics-guided
  human motion diffusion model,'' in {\em IEEE/CVF International Conference on
  Computer Vision (ICCV)}, 2023.

\bibitem{oreshkin2023motion}
B.~N. Oreshkin, A.~Valkanas, F.~G. Harvey, L.-S. M{\'e}nard, F.~Bocquelet, and
  M.~J. Coates, ``Motion in-betweening via deep delta-interpolator,'' {\em IEEE
  Transactions on Visualization and Computer Graphics}, 2023.

\bibitem{petrovich23tmr}
M.~Petrovich, M.~J. Black, and G.~Varol, ``{TMR}: Text-to-motion retrieval
  using contrastive {3D} human motion synthesis,'' in {\em International
  Conference on Computer Vision ({ICCV})}, 2023.

\bibitem{chicodataset}
A.~Sampieri, G.~M.~D. di~Melendugno, A.~Avogaro, F.~Cunico, F.~Setti,
  G.~Skenderi, M.~Cristani, and F.~Galasso, ``Pose forecasting in industrial
  human-robot collaboration,'' in {\em European Conference on Computer Vision},
  pp.~51--69, Springer, 2022.

\bibitem{stsgcn}
T.~Sofianos, A.~Sampieri, L.~Franco, and F.~Galasso, ``Space-time-separable
  graph convolutional network for pose forecasting,'' in {\em IEEE/CVF
  International Conference on Computer Vision}, pp.~11209--11218, 2021.

\end{thebibliography}

\end{document}